%% file: fmlp-projection-npl-preprint.tex
\documentclass[numreferences]{kluwer}
\usepackage[latin1]{inputenc}
\usepackage{amsmath}

\newcommand{\R}[1]{\ensuremath{\dR^{#1}}}
\newcommand{\N}[1]{\ensuremath{\dN^{#1}}}
\newcommand{\E}[1]{\mathbf{E}\left[{#1}\right]}
\newcommand{\error}{\mathcal{C}}
\usepackage{url}
\newcommand{\di}{\mathrm{d}}
\newdisplay{definition}{Definition}
\newdisplay{theorem}{Theorem}

\begin{document}
\sloppy
\begin{frontmatter}
  
\end{frontmatter}
\begin{article}
  \begin{opening}
\title{Theoretical Properties of Projection Based Multilayer Perceptrons with
  Functional Inputs\protect\footnote{Published in Neural Processing Letters
    (Volume 23, Number 1, February 2006, pages 55--70). The original
    publication is available at \url{www.spingerlink.com}. DOI: \url{http://dx.doi.org/10.1007/s11063-005-3100-2}}}    
\author{Fabrice \surname{Rossi}\email{Fabrice.Rossi@inria.fr}}
\institute{Projet AxIS, INRIA, Domaine de Voluceau, Rocquencourt,
  B.P. 105, 78153 Le Chesnay Cedex, France}
\author{Brieuc \surname{Conan-Guez}\email{Brieuc.Conan-guez@iut.univ-metz.fr}}
\institute{LITA EA3097, Université de Metz, Ile du Saulcy, F-57045 Metz,
France}
\date{\today}
\begin{abstract}
Many real world data are sampled functions. As shown by Functional Data
Analysis (FDA) methods, spectra, time series, images, gesture recognition
data, etc. can be processed more efficiently if their functional nature is
taken into account during the data analysis process. This is done by extending
standard data analysis methods so that they can apply to functional inputs. A
general way to achieve this goal is to compute projections of the functional
data onto a finite dimensional sub-space of the functional space. The
coordinates of the data on a basis of this sub-space provide standard vector
representations of the functions. The obtained vectors can be processed by any
standard method. 

In \cite{RossiEtAl05Neurocomputing}, this general approach has been used
to define projection based Multilayer Perceptrons (MLPs) with functional
inputs. We study in this paper important theoretical properties of the
proposed model. We show in particular that MLPs with functional inputs are
universal approximators: they can approximate to arbitrary accuracy any
continuous mapping from a compact sub-space of a functional space to \R{}.
Moreover, we provide a consistency result that shows that any mapping from a
functional space to \R{} can be learned thanks to examples by a projection
based MLP: the generalization mean square error of the MLP decreases to the
smallest possible mean square error on the data when the number of examples
goes to infinity.
\end{abstract}
\end{opening}

\section{Introduction}
In many practical situations, input data are in fact sampled functions rather
than standard high dimensional vectors. This is the case for instance in
spectrometry: a discretized spectrum is obtained by measuring the
transmittance or the reflectance of an object at different wavelengths. Modern
spectrometers can produce very high resolution spectra, with a thousand of
observations for each spectrum. 

Another general example of sampled functions is given by time series. Indeed, a
time series is a mapping from a time period to a observation range, for
instance the hourly temperature at a weather station over one month. More
complex examples can be found in meteorology, for instance rainfall maps,
i.e., functions that map geographical coordinates and date to the daily rain
level observed at the specified position and date. 

Functional Data Analysis (FDA) \cite{BesseRamsay1986,RamseyDalzell1991} is a
general methodology targeted at data that are better described as functions
than as vectors. The main idea is to take advantage of the functional nature
of the data to design better data analysis methods than the ones constructed
thanks to a vector model. For a comprehensive introduction to FDA methods we
refer the reader to \cite{RamseySilverman97} in which extensions of classical
data analysis tools to functional data, developed since pioneering works such
as \cite{Deville74} and \cite{DauxoisPousse76}, are precisely described. 

The simplest case of FDA corresponds to a situation in which all considered
functions are discretized at the same points. More precisely, if we consider
$n$ functions $g^1,\ldots,g^n$ and $m$ sampling points $x_1,\ldots,x_m$, we
obtain $n$ vector from $\R{m}$, $(g^i(x_1),\ldots,g^i(x_m))$. While direct
comparison between vectors remains possible, this type of data suffers from
two drawbacks: high dimension vectors and high correlation between variables.
\cite{RamseySilverman97} focuses on this situation and provides solutions that
explicitly use the underlying functions $(g^i)_{1\leq i\leq n}$. The general
methodology uses the fact that most multivariate data analysis methods are
based on scalar products and/or distance calculations which can be easily
translated from a finite dimensional space to a functional space. A simple
example is given by linear regression: if we want to predict a target variable
in \R{}, $y^i$ with a linear model on $g^i$, the classical model on the
discretized function tries to model $y^i$ as:
\begin{equation}\label{equationdiscretizedLinearModel}
y^i = w_0 + \sum_{j=1}^mw_j g^i(x_j) + \varepsilon^i
\end{equation}
A functional version is given by
\cite{RamseyDalzell1991,HastieMallows1993,CardotFerratySarda1999}: 
\begin{equation}\label{equationFunctionalLinearModel}
y^i = w_0 + \int w(x)g^i(x)\di x + \varepsilon^i,
\end{equation}
in which most numerical parameters of the model ($w_1,\ldots,w_m$) have been
replaced by an unique functional parameter. A simple yet powerful idea to
implement the functional version of the model is to estimate each $g^i$ thanks
to the corresponding vector $(g^i(x_1),\ldots,g^i(x_m))$ and then to work on
the approximated function. Classical solutions are based on spline
approximations of both $g^i$ and $w$ (see
\cite{HastieMallows1993,MarxEilers1996,CardotFerratySarda2002} for instance).
They solve the variable correlation problem by reducing the effective
dimension of the functional parameter thanks to regularity assumptions (e.g.
bounded second derivative).

A very interesting side effect of estimating $g^i$ thanks to its discretized
version is to allow processing of irregularly sampled functions that are quite
common in many applications, especially medical ones (see
e.g.
\cite{BrumbackRice1998,HooverRiceWuYang1998,JamesHastieSugar2000,RiceWu2001}).
Group smoothing techniques have been developed for these types of data: rather
than estimating each $g^i$ independently, one can try to optimize the global
representation of all examples, either by EM like methods
\cite{JamesHastieSugar2000,RiceWu2001} or using hybrid splines and
cross-validation \cite{BesseCardotFerraty1997}. Moreover, functional
transformations (such as derivative calculation, see
\cite{FerratyVieu2002CSDA,FerratyVieu2002CS}) can be  
performed on the representation. It is therefore obvious that the functional
view of high dimensional data gives much more possibilities than the bare
multivariate analysis. 

Many classical data analysis tools have been adapted to functional
data. Principal Component Analysis was the first method studied in a
functional framework by \cite{Deville74} and \cite{DauxoisPousse76} (see
also \cite{DauxoisPousseRomain82,RamseySilverman97} and
\cite{JamesHastieSugar2000}). Other linear 
methods have been studied more recently, such as Canonical Correlation
Analysis \cite{LeurgansMoyeedSilverman1993}, linear discriminant analysis
\cite{HastieBujaTibshirani1995,JamesHastie2001} and linear regression (as
presented above
\cite{RamseyDalzell1991,HastieMallows1993,CardotFerratySarda1999}). Non linear
models such as generalized linear models \cite{James2002}, slice inverse
regression \cite{FerreYao2003SIR} and non-parametric kernel based estimation
\cite{FerratyVieu2002CSDA,FerratyVieu2002CS} have also been reformulated to
work on functional data. Unsupervised classification of functions has also been
studied, as a quantization problem in \cite{LuschgyPages2002} and
more traditionally with k-means like approaches \cite{Abraham2000} or mixture
models \cite{JamesSugar2002}. Additional references and discussions about
functional data analysis can be found in \cite{RamseySilverman97}. 

Neural models have been recently adapted to functional data (see
\cite{RossiConanGuez05NeuralNetworks,RossiEtAl05Neurocomputing,RossiConanGuezElGolliESANN2004SOMFunc,DelannayEtAlESANN2004RbfFunc,RossiConanESANN2004FMLPMissing}).
Building on extensions of multilayer perceptrons (MLPs) to arbitrary inputs
studied in \cite{Sandberg1994,Sandberg1996,SandbergXu1996,Stinchcombe99}, we
have proposed in \cite{RossiConanGuez05NeuralNetworks} a functional multilayer
perceptron (FMLP) based on approximate calculation of some integrals. While
this model has interesting theoretical properties (cf
\cite{RossiConanGuez05NeuralNetworks,RossiConanGuez05CRASConsistance}) and
gives very satisfactory results on real world benchmarks, it suffers from the
need of a specialized implementation and from long training times. In
\cite{RossiEtAl05Neurocomputing}, we have proposed another functional MLP
based on projection operators. This method has some advantages over the one
studied in \cite{RossiConanGuez05NeuralNetworks}, especially because the
projections can be implemented as a pre-processing step that transforms
functions into adapted vector representations. The vectors obtained like this
are then processed by a standard neural model. We have shown in
\cite{RossiEtAl05Neurocomputing} that this functional model performs very well
on real world data. However, this illustration was only experimental.

In this paper, we study theoretically the capabilities of projection based
functional MLPs. We first recall in section \ref{sectionFMLP} the definition
of the functional multilayer perceptron and its projection based
implementation. In section \ref{sectionUniversalApproximation}, we show that
any continuous function from a compact sub-space of a functional space to \R{}
can be approximated arbitrarily well by projection based FMLPs, which are
therefore universal approximators. In section \ref{sectionConsistency}, we
show that functional MLPs can learn arbitrary mappings from a functional space
to \R{}. More precisely, we show that the asymptotic generalization error of
functional MLPs converges to the minimum possible error, provided the training
is done properly. Proofs are gathered in section \ref{sectionProofs}.

\section{Multilayer perceptrons with functional inputs}\label{sectionFMLP}
\subsection{Introduction}
In this section, we recall the definition of functional multilayer perceptrons
given in \cite{RossiEtAl05Neurocomputing}. We focus on regular functions. More
precisely, we denote $\mu$ a $\sigma$-finite positive Borel measure defined on
$\R{p}$ and $L^2(\mu)$ the space of measurable real valued
functions\footnote{More precisely, $L^2(\mu)$ contains equivalence classes of
  functions that  differ only on a $\mu$-negligible set.} defined
on $\R{p}$ and such that $\int f^2\di\mu<\infty$. $L^2(\mu)$ is a Hilbert
space equipped with its natural inner product $\langle f,g\rangle = \int
fg\di\mu$ (we denote $\|f\|_2=\sqrt{\langle f,f\rangle}$). 

To avoid cumbersome notations, this paper is restricted to data described by a
single function valued variable. However, the results can be easily extended to
the case of data described by several functional variables. We also restrict
ourselves to one real valued output, but results are also valid for vector
valued output.

\subsection{Theoretical model}
As recalled in the introduction and explained in \cite{RamseySilverman97},
many data analysis methods are based on the Hilbert structure of the input
space rather than on its finite dimension. Using this idea,
\cite{RossiEtAl05Neurocomputing} defines multilayer perceptrons with
functional inputs, as recalled here.

A multilayer perceptron (MLP) consists in neurons that perform very simple
calculations. Given an input $x\in\R{p}$, the output of a neuron is
\begin{equation}
T\left(\beta_0+\sum_{i=1}^p\beta_ix_i\right), 
\end{equation}
where $x_i$ is the $i$-th coordinate of $x$, $T$ is an activation function
from \R{} to \R{}, and $\beta_0,\ldots,\beta_p$ are numerical parameters (the
weights of the neuron).

The sum $\sum_{i=1}^p\beta_ix_i$ is in fact the inner product in $\R{p}$
between $x$ and $(\beta_1,\ldots,\beta_p)$. As proposed in
\cite{RossiConanGuez05NeuralNetworks,RossiEtAl05Neurocomputing}, a functional
neuron can be defined thanks to the inner product in $L^2(\mu)$. Given an
input $g\in L^2(\mu)$, the output of a functional neuron is 
\begin{equation}
T(\beta_0+\langle w,g\rangle)=T\left(\beta_0+\int wg\di\mu\right), 
\end{equation}
where $w$ is a function from $L^2(\mu)$, the ``weight function''. This
functional neuron is in fact a special case of neurons with arbitrary input
spaces defined in previous theoretical works
\cite{Sandberg1994,Sandberg1996,SandbergXu1996,Stinchcombe99}. 

As the output of a generalized neuron is a numerical value, we need such
neurons only in the first layer of the MLP. Indeed, the second layer uses only
outputs from the first layer which are real numbers and therefore consists in
numerical neurons. For example, a single hidden layer perceptron with an
unique output neuron maps a functional input $g$ to
\begin{equation}
H(g)= \sum_{l=1}^L a_l T\left(\beta_{0l}+\int w_l g \di \mu \right),
\end{equation}
where $L$ denotes the number of hidden (functional) neurons and
$a_1,\ldots,a_L$ are real valued connexion weights of the output neuron (it
has a linear activation function).

\subsection{Projection}\label{subsectionProjection}
While the model presented in the previous section is a simple generalization of
its numerical counterpart, it cannot be used in practice, as only a
limited class of functions can be easily manipulated on a computer. Those
functions are obtained as combinations (sum, product, composition, etc.) of
elementary functions: polynomial functions, trigonometric functions, etc.

In order to solve this problem, FDA methods rely in general on projections.
Let us indeed consider a finite $p$-dimensional subspace of $L^2(\mu)$,
denoted $V_p$. The main principle of projection based FDA methods is to
constrain all manipulated functions to belong to $V_p$ rather than to
$L^2(\mu)$. This constraint is implemented thanks to an orthogonal projection
on $V_p$. More precisely, let us denote $\Pi_p$ the orthogonal projection
operator on $V_p$. Given an arbitrary input function $g$, the
output of a functional neuron constructed thanks to $V_p$ is given by
\begin{equation}
T\left(\beta_0+\int \Pi_p(w)\Pi_p(g)\di\mu\right).
\end{equation}
The main advantage of using $V_p$ is that it can be obtained as the vector
space spanned by ``computer friendly'' functions, that is, functions that are
easy to evaluate on a computer. One possibility consists in using a Hilbert
basis of $L^2(\mu)$, that is a complete orthonormal system
$(\phi_k)_{k\in\N{*}}$. Useful examples include
wavelets and trigonometric functions. Then $V_p$ is defined as the vector
space spanned by $(\phi_k)_{1\leq k\leq p}$. 

Another possibility consists in using spline spaces, that is vector spaces of
piecewise polynomial functions, or more generally, specific $V_p$ that have
been chosen because $\Pi_p$ is easy to calculate and functions in $V_p$ are
easy to manipulate.

On the theoretical point of view and in the general case, $V_p$ is given by an
orthonormal basis $(\phi_{p,k})_{1\leq k\leq p}$. This basis allows to identify
$V_p$ with $\R{p}$. We denote $\pi_p$ the coordinate map, that is the function
from $L^2(\mu)$ to \R{p} that maps $g$ to the coordinates of $\Pi_p(g)$ on the
basis $(\phi_{p,k})_{1\leq k\leq p}$, i.e., to a vector in \R{p} such that
$\Pi_p(g)=\sum_{k=1}^p\pi_p(g)_k\phi_{p,k}$. We have:
\begin{equation}
\int \Pi_p(w)\Pi_p(g)\di\mu=\sum_{k=1}^p\pi_p(w)_k\pi_p(g)_k.
\end{equation}
This shows, as explained in \cite{RossiEtAl05Neurocomputing}, that the
projection approach corresponds to a pre-processing step that transforms
functional inputs into finite dimensional inputs. A simple way to implement a
projection based functional MLP consists in using a standard MLP to which the
$p$ coordinates of the projected functions are submitted (the MLP uses
therefore standard vector inputs in \R{p}). The resulting model gives exactly
the same output as a functional MLP build for functional inputs in $V_p$.

\section{Universal approximation}\label{sectionUniversalApproximation}
\subsection{Definition}
This section is dedicated to the approximation capabilities of the functional
MLP described in the previous section. We first recall a definition of
universal approximation. 

If $A$ and $B$ are two topological spaces, we denote $C(A,B)$ the set of
continuous functions from $A$ to $B$. 
\begin{definition}
Let $X$ be a topological space and $\mathcal{B}$ be a set of continuous
functions from $X$ to \R{}. We say that $\mathcal{B}$ has the universal
approximation property for $X$ if for any compact subset of $X$, $K$,
$\mathcal{B}$ is dense in $C(K,\R{})$ for the uniform norm. 
\end{definition}
In other words, if $\mathcal{B}$ has the universal approximation property for
$X$, for any compact subset $K$, any continuous function $f$ from $K$ to \R{},
and any requested precision $\epsilon>0$, there is $g\in\mathcal{B}$ such that
$\sup_{x\in K}|f(x)-g(x)|=\|f-g\|_{\infty}<\epsilon$. 

\subsection{Projection and universal approximation}
When functions are processed thanks to a projection, approximation
capabilities depend both on the neural model and on the projection. It is
quite obvious that universal approximation cannot be reached if MLPs are
constrained to work on a fixed $V_p$ subset. Indeed, most of the functions in
$L^2(\mu)$ are very poorly approximated by their projections on $V_p$ for a
fixed set of functions $(\phi_k)_{1\leq k\leq p}$. Therefore, the neural
models have not enough information on their actual inputs to provide 
meaningful outputs. To solve this problem, we need to consider more and more
precise projections.
\begin{definition}
  Let us consider a sequence of functions from $L^2(\mu)$
  $(\phi_{p,k})_{p\in\N{*},1\leq k\leq p}$ such that for each $p$,
  $(\phi_{p,k})_{1\leq k\leq p}$ is an orthonormal system. We denote $V_p$ the
  subspace of $L^2(\mu)$ spanned by $(\phi_{p,k})_{1\leq k\leq p}$ and $\Pi_p$
  the orthogonal projection operator on $V_p$.
  
  Let $\mathcal{G}$ be a subset of $L^2(\mu)$. The sequence $(\Pi_p)_{p \in
    \N{*}}$ (and the corresponding sequence of functions) is said to have the
  point-wise approximation property for $\mathcal{G}$ if $\Pi_p$ converges to
  $Id_{\mathcal{G}}$ on $\mathcal{G}$ for the point-wise convergence: for all
  $g \in \mathcal{G}$, $\lim_{p\rightarrow\infty}\|\Pi_p(g)-g\|_2$=0.
\end{definition}
A simple example of sequence with the point-wise approximation property for
$L^2(\mu)$ is given by any Hilbert basis $(\phi_k)_{k\in\N{*}}$ of this
space. Indeed, any function $g$ in $L^2(\mu)$ has a series expansion
$g=\sum_{k=1}^\infty g_k\phi_k$. Therefore, the sequence defined by
$\phi_{p,k}=\phi_k$ has obviously the point-wise approximation property. 

Thanks to those increasingly accurate projections, we can construct a set of
MLP based functions with the universal approximation property for $L^2(\mu)$.
\begin{theorem}\label{theoremApproxUniv}
Let $T$ be a continuous non polynomial function from \R{} to \R{} and let
$(\phi_{p,k})_{p\in\N{*},1\leq k\leq p}$ be a sequence of functions from 
$L^2(\mu)$ with the point-wise approximation property for $L^2(\mu)$. Let us
denote $\mathcal{S}(T,(\phi_{p,k})_{p\in\N{*},1\leq k\leq p})$ the set of
functions from $L^2(\mu)$ to \R{} of the form 
\[
g\mapsto \sum_{l=1}^La_lT\left(\beta_{l0}+\sum_{k=1}^p\beta_{lk}\pi_{p}(g)_k\right),
\]
where $L\in\N{*}$, $p\in\N{*}$, $\beta_{lk}\in\R{}$ and $a_l\in\R{}$ ($\pi_p$
is the coordinate map defined in section \ref{subsectionProjection}). 

Then $\mathcal{S}(T,(\phi_{p,k})_{p\in\N{*},1\leq k\leq p})$ has the universal
approximation property for $L^2(\mu)$. 
\end{theorem}

\subsection{Relation to previous works}
A lot of work has been done on the topic of universal approximation properties
of multilayer perceptrons (see, e.g.,
\cite{Stinchcombe99,Pinkus1999Approximation} for reviews). For functional
inputs, pioneering work can be found in \cite{ChenChen1993FMLP}. This paper
proves that single hidden layer perceptrons with functional inputs have the
universal approximation property for $C([a,b],\R{})$ and $L^p([a,b])$. Those
results are based either on the exact calculation of some specific
integrals (for $L^p([a,b])$) or on a vector representation of functions based
on an evaluation map (for $C([a,b],\R{})$): $g$ is replaced by
$(g(x_1),\ldots,g(x_n))$. Those results have been improved in more recent
papers \cite{ChenChen1995b,Chen1998}. 

Other pioneering work can be found in \cite{Sandberg1994}: this paper shows
that some specific feed-forward architecture with functional inputs has the
universal approximation property. This result relies on perfect calculation of
inner products.  Generalizations of this result can be found in
\cite{SandbergXu1996,Stinchcombe99}.

Finally, \cite{ChenChen1995RBF} studies a projection based approach for Radial
Basis Function Network and \cite{Sandberg1996} studies the approximate
realization of the model proposed in \cite{Sandberg1994} thanks to projection.
Both works are related to the model proposed in the present paper. The novelty
of our approach consists in allowing complex projection methods whereas
\cite{ChenChen1995RBF,Sandberg1996} are limited to truncated basis
representation. The complex projection methods covered by Theorem
\ref{theoremApproxUniv}, especially those based on spline approximations, have
been used successfully in \cite{RossiEtAl05Neurocomputing} for real world
data.

\section{Consistency}\label{sectionConsistency}
\subsection{Introduction}
While universal approximation is an important property, it is not sufficient
to ensure that the considered model can be used with success for some machine
learning task. Another problem must be assessed: is it possible to design, from
a finite set of examples, a functional MLP such that when the number of
examples goes to infinity, the FMLP provides a more and more accurate
approximation of the underlying relationship between the input functions and
the numerical outputs? This question (the ``learnability'') has been studied in
details in the case of numerical MLP, see e.g.
\cite{White1990NonParametric,Barron1994,LugosiZeger1995}.

To give a precise mathematical translation of this question, we introduce the
following notations (we follow \cite{LugosiZeger1995}). We denote $(G,Y)$ a
pair of random variables, defined on probability space $ (\Omega
,\mathcal{M},P) $, that take their values from $L^2(\mu)$ and \R{},
respectively. Our goal is to predict the value of $Y$ given $G$. To assess the
quality of this prediction, we need an error measure. In this paper, we use
the root mean square error, but any $L_p$-error could be used\footnote{There
  is no relation between the functional input space $L^2(\mu)$ and the use of
  the mean square error.}.  Given a
function $h$ from $L^2(\mu)$ to \R{}, the root mean square prediction error is
defined as
\begin{equation}
  \label{eqMeanSquareError}
\error(h)=\E{(h(G)-Y)^2}^{\frac{1}{2}},
\end{equation}
where $\E{.}$ denotes the expectation. If we assume that $\E{|Y|^2}<\infty$,
then $\error$ is minimized by the conditional expectation of $Y$ given $G$,
i.e., by $h(g)=\E{Y|G=g}$. We denote $\error^*$ the minimal root mean square
error, i.e.
\begin{equation}
  \label{eqMinimalRMS}
\error^*=\inf_{h}\error(h)=\E{(\E{Y|G}-Y)^2}^{\frac{1}{2}}.
\end{equation}
We have no information about the distribution of $(G,Y)$, except for $n$
independent, identically distributed (i.i.d.) copies of $(G,Y)$, 
\[
D_n=((G^1,Y^1),\ldots,(G^n,Y^n)).
\]
Using this data set, we can build a prediction model $h_n$ (from $L^2(\mu)$
to \R{}). The model depends on $D_n$ and its performances are given by the
following random variable 
\begin{equation}
  \label{eqPerfModel}
  \error(h_n)=\E{(h_n(G)-Y)^2|D_n}^{\frac{1}{2}}.
\end{equation}
A sequence of prediction models $(h_n)_{n\in\N{*}}$ is \textbf{universally
  consistent} (see \cite{LugosiZeger1995}) if $\error(h_n)$ converges almost
surely to $\error^*$, for any distribution $(G,Y)$ satisfying
$\E{|Y|^2}<\infty$. The intuitive interpretation of this condition is that
given enough data (when $n$ goes to infinity), the root mean square error of
$h_n$ will be arbitrarily close to the best possible root mean square
error: we are indeed \emph{learning} the relationship between $Y$ and $G$ from
examples. Another way to look at the condition is to rewrite it into the
following equivalent condition:
\begin{equation}
\E{(h_n(G)-\E{Y|G})^2|D_n}^{\frac{1}{2}}\xrightarrow[n\rightarrow\infty]{} 0\
a.s. 
\end{equation}
This condition means that $h_n(G)$ is arbitrarily close to $\E{Y|G}$ for the
mean square error.

\subsection{Projection and consistency}
In this section, we restrict the projection approach to the simple case of
sequences of projection spaces constructed thanks to a Hilbert basis of the
functional space. More precisely, we assume given $(\phi_p)_{p\in\N{*}}$ a
Hilbert basis of $L^2(\mu)$. We denote $V_p$ the sub-vector space spanned by
$(\phi_k)_{1\leq k\leq p}$ and define $\pi_p$ as in section
\ref{subsectionProjection}. In order to build a consistent learning method
based on projection on $V_p$ spaces, we need to adapt the expressive power of
the candidate neural networks to the size of the learning set (i.e., to $n$).
Rather than choosing an arbitrary single hidden layer perceptron, we restrict
the search to some classes of such perceptrons. More precisely, given
$(L_n)_{n\in\N{*}}$ a sequence of integers and $(\alpha_n)_{n\in\N{*}}$ a
sequence of positive real values, we define $\mathcal{H}_{np}$, a sequence of
single hidden layer functional perceptron classes, by:
\begin{equation} \label{eqFMLPClasses}
  \begin{split}
    \mathcal{H}_{np}=\Biggl\{&h\in C(L^2(\mu),\R{})\Biggr|
    \\
&
    h(g)=\sum_{l=1}^{L_n}a_lT\left(\beta_{l0}+\sum_{k=1}^p\beta_{lk}\pi_{p}(g)_k\right),
    \sum_{k=1}^{L_n}|a_l|\leq \alpha_n\Biggr\}.
\end{split}
\end{equation}
In those classes, $L_n$ and $\alpha_n$ provide a type of regularization by
adapting the number of hidden neurons (thanks to $L_n$) and the magnitude
of the weights of the output layer (thanks to $\alpha_n$) to the size of the
learning set. A consistent sequence of models will be obtained by choosing the
best single hidden layer perceptron in $\mathcal{H}_{np}$, according to the
empirical error (see  Theorem \ref{theoremConsistency}). 

To obtain consistency, we need some technical hypotheses:
\begin{itemize}
\item[(H-1)] $T$ is a function from \R{} to $[0,1]$, monotone non
decreasing, with $\lim_{x\rightarrow \infty}T(x)=1$ and $\lim_{x\rightarrow
  -\infty}T(x)=0$;
\item[(H-2)] $(L_n)_{n\in\N{*}}$ and $(\alpha_n)_{n\in\N{*}}$ are such that
  \begin{gather*}
    \lim_{n\rightarrow\infty} L_n=\infty\\
    \lim_{n\rightarrow\infty} \alpha_n=\infty;\\
  \end{gather*}
\item[(H-3)] $(L_n)_{n\in\N{*}}$ and $(\alpha_n)_{n\in\N{*}}$ are such that
\[
    \lim_{n\rightarrow\infty}\frac{L_n\alpha_n^{4}\log(L_n\alpha_n)}{n}=0,
\]
  and such that there is $\delta>0$ such that
\[
\lim_{n\rightarrow\infty}\frac{\alpha_n^{4}}{n^{1-\delta}}=0.
\]
\end{itemize}
Hypothesis (H-1) corresponds to a standard requirement for activation
functions of multilayer perceptrons. It is fulfilled for instance by
$T(x)=1/(1+e^{-x})$. 

Hypothesis (H-2) ensure that the expressive power of the considered classes is
not limited asymptotically, as the regularization vanishes asymptotically. 

Hypothesis (H-3) corresponds the regularization. The constraints come from
\cite{LugosiZeger1995} (stronger constraints were used in
\cite{White1990NonParametric}). They control the way the expressive power of
$\mathcal{H}_{np}$ grows with $n$. 

Some possible choices for $L_n$ and $\alpha_n$
include $L_n=\lceil\log n\rceil$ (where $\lceil x\rceil$ denotes the smallest
integer greater or equal to $x$) and $\alpha_n=n^{\frac{1}{8}}$.

Under those hypotheses, we have the following consistency result.
\begin{theorem}\label{theoremConsistency}
Let $h_{np}$ be a function that minimizes the empirical mean square error
in $\mathcal{H}_{np}$, i.e. such that
\[
\frac{1}{n}\sum_{i=1}^n(h_{np}(G^i)-Y^i)^2\leq \frac{1}{n}\sum_{i=1}^n(h(G^i)-Y^i)^2,
\]
for all $h\in \mathcal{H}_{np}$. 

Under hypotheses (H-1), (H-2) and (H-3), we have
\[
\lim_{p\rightarrow\infty}\lim_{n\rightarrow\infty}\error(h_{np})=\error^*\ [a.s.],
\]
for all distributions of $(G,Y)$ such that $\E{|Y|^2}<\infty$. 
\end{theorem}
The theorem means that a functional MLP $h_{np}$ that minimizes its mean
square error on a training set with $n$ examples provides a more and more
accurate approximation of $\E{Y|G}$ when $n$ goes to infinity. The theorem
provides some rules on $L_n$ and $\alpha_n$ that allow to avoid over-fitting and
guarantee good generalization. The only limitation of this result comes from
the sequential limit: the theorem does not provide guidelines to link $p$ to
$n$.

It should be noted that the theorem could be adapted to any model that is
universally consistent in finite dimension. 

\section{Conclusion}
We have demonstrated in this paper two important results for projection based
functional multilayer perceptrons: they have the universal approximation
property and they can learn arbitrary mapping. Thanks to the representation of
the studied functions through projection, we adapt the strong results
available for standard numerical MLPs to functional MLPs. This gives a
satisfactory theoretical backing to the method proposed in
\cite{RossiEtAl05Neurocomputing}. 

However, some questions remain open, especially if we want to fully justify
the method illustrated in \cite{RossiEtAl05Neurocomputing}. The more important
point is the choice of the projection quality, i.e., of $V_p$. In
\cite{RossiEtAl05Neurocomputing}, it was determined thanks to the input data
alone. The goal was to limit the distortion between $\Pi_p(g)$ and
$g$. Further theoretical investigation of this method is needed.

Another possibility is to use a split sample or a re-sampling technique to
choose an optimal $V_p$, as we did in
\cite{RossiConanGuez05NeuralNetworks}. In practice, this introduces a huge
computational load, without much practical gain. However, models constructed
like this are universally consistent in the case of classification
\cite{BiauEtAl2005FunClassif}. 

In practice, some very good results have been obtained in
\cite{FerreVilla04SIRNN} thanks to an automatic construction of $V_p$ based on
a functional version of the Slice Inverse Regression. While the authors
provide important theoretical results, the consistency of this method remains
an open question.

Finally, the second more important open question is related to the very nature
of functional data: in practice, functional data are always given as finite
sets of (input, output) pairs. As a consequence, projected functions cannot be
exactly computed and are replaced by approximations (see
\cite{RossiEtAl05Neurocomputing} for details). The effects of this
approximation on the capabilities of functional MLP have been partially
studied in \cite{ConanRossiICANN2002} but the consistency of models
constructed thanks to approximate values is not yet established. 

\section{Proofs}\label{sectionProofs}
\subsection{Theorem \ref{theoremApproxUniv}}
This theorem is based on results given in \cite{Stinchcombe99} for MLPs with
arbitrary inputs. 

Let us consider a compact subset of $L^2(\mu)$, $K$. We want to approximate
functions in $C(K,\R{})$ by functions in
$\mathcal{S}(T,(\phi_{p,k})_{p\in\N{*},1\leq k\leq p})$.

\subsubsection{Step one}
As a first step, we prove that the sequence of operator $(\Pi_p)_{p \in
  \N{*}}$ converges to $Id_K$ uniformly on $K$, i.e. for $\eta>0$, there is
$P$ such that for each $p\geq P$ and for each $g \in K$, $\| \Pi_p(g) - g
\|_2< \eta$ ($P$ does not depend on $g$). 

Let us consider $g_0 \in K$, and $K(g_0,r)=B(g_0,r)\cap K$ neighborhood of
$g_0$ in $K$, where $B(g_0,r)$ denotes the open ball of radius $r$ centered on
$g_0$. As $(\Pi_p)_{p \in \N{*}}$ has the point-wise approximation property,
there is $P_0$ such that for each $p_0\geq P_0$, $\| \Pi_{p_0} (g_0) - g_0
\|_{2}< \eta /2$.  For each $g \in K(g_0,r)$, we have $\| \Pi_{p_0}(g) - g
\|_{2} \leq \| \Pi_{p_0}(g) - \Pi_{p_0}(g_0)\|_{2} +\| \Pi_{p_0}(g_0) - g_0
\|_{2} +\| g_0 - g \|_{2}$.  As requested above, the middle term is smaller
than $\eta/2$.  As $\Pi_{p_0}$ is Lipschitz continuous (the Lipschitz constant
is 1), $\| \Pi_{p_0}(g)-\Pi_{p_0}(g_0)\|_{2} \leq \| g - g_0 \|_{2}$.
Therefore, $\| \Pi_{p_0}(g) - g \|_{2} \leq \eta/2 + 2\| g - g_0 \|_{2}$. As a
consequence, $\forall g\in K(g_0,\eta/4)$, $\|\Pi_{p_0}(g) - g \|_{2} <
\eta$. As $K$ is compact, it is covered by a finite number of $B(g_i,\eta/4)$
(and therefore of $K(g_i,\eta/4)$). We consider $P=\max P_i$, which allows to
conclude.

\subsubsection{Step two}
Let us now denote $\mathcal{S}(T,L^2(\mu))$ the set of functions
form $L^2(\mu)$ to \R{} of the form  
\begin{equation}\label{ieqDirectFMLP}
g\mapsto \sum_{l=1}^La_lT\left(\beta_{l0}+\langle w_l,g\rangle\right),
\end{equation}
where $l\in\N{*}$, $p\in\N{*}$, $\beta_{l0}\in\R{}$ and $w_l\in
L^2(\mu)$. Then, $\mathcal{S}(T,L^2(\mu))$ has the universal approximation
property for $L^2(\mu)$. 

Indeed, Corollary 5.1.2 of \cite{Stinchcombe99} can be applied, as its
conditions are fulfilled:
\begin{itemize}
\item $L^2(\mu)$ is locally convex and is isometric to its topological dual;
\item as $T$ is continuous and non-polynomial, single hidden layer perceptrons
  using $T$ as their activation function have the universal approximation
  property for \R{} (see \cite{Pinkus1999Approximation} for instance).
\end{itemize}

\subsubsection{Step three}
Let us now consider a continuous function $F$ from $K$ to \R{} and let
$\epsilon>0$ be an arbitrary precision. 

According to step two, there is $H\in \mathcal{S}(T,L^2(\mu))$, given by
equation \ref{eqDirectFMLP}, such that for all $g\in K$, $|H(g)-F(g)|<
\frac{\epsilon}{2}$.

As $H$ is continuous on $L^2(\mu)$, for each $g \in K$ there is $\eta(g)>0$
such that for each $f \in B(g,\eta(g))$, we have $|H(g)-H(f)| \leq
\frac{\epsilon}{4}$. As $K$ is compact, it is covered by a finite number of
the balls $\left(B\left(g_i,\frac{\eta(g_i)}{2}\right)\right)_{1\leq i\leq
  N}$. We denote $\eta=\min_{1\leq i\leq N}\eta(g_i)$.

According to Step one of the proof, there is $p$ such that for all $g\in K$,
$\|\Pi_p(g)-g\|_{2}<\frac{\eta}{2}$. There is $i$ such that $g$ falls in
$B\left(g_i,\frac{\eta(g_i)}{2}\right)$ and we have 
\[
\|\Pi_p(g)-g_i\|_{2}\leq \|\Pi_p(g)-g\|_{2}+\|g-g_i\|_{2}<\eta(g_i),
\]
which implies
\[
|H(\Pi_p(g))-H(g)|\leq |H(\Pi_p(g))-H(g_i)|+|H(g_i)-H(g)|<\frac{\epsilon}{2},
\]
by using twice the continuity of $H$ at $g_i$. We have therefore
\[
|H(\Pi_p(g))-F(g)|<\epsilon.
\]
To conclude, we note that $\langle w_l,\Pi_p(g)\rangle=\langle
\Pi_p(w_l),\Pi_p(g)\rangle=\sum_{k=1}^p\pi_p(w_l)_k\pi_p(g)_k$, which means
that $H\circ \Pi_p$ belongs to $\mathcal{S}(T,(\phi_{p,k})_{p\in\N{*},1\leq
  k\leq p})$.

\subsection{Theorem \ref{theoremConsistency}}
Theorem \ref{theoremConsistency} is based on theorem 3 from
\cite{LugosiZeger1995}.  The latter applies to standard MLPs with inputs in
\R{d} and provides universal consistency.

\subsubsection{Step one}
To use theorem 3 from \cite{LugosiZeger1995}, we need to introduce additional
notations. We denote $G_p=\pi_p(G)$. As $\pi_p$ is continuous, $G_p$ is a
random variable that takes values from \R{p}. We denote $G^i_p=\pi_p(G^i)$.
Obviously, for any $p$, $D^p_n=((G^1_p,Y^1),\ldots,(G^n_p,Y^n))$ consists in
$n$ i.i.d. copies of $(G_p,Y)$.

If $f_n$ is a measurable function from $\R{p}$ to \R{} constructed thanks to
$D^p_n$, we denote 
\[
\error_p(f_n)=\E{(f_n(G_p)-Y)^2|D^p_n}^{\frac{1}{2}}.
\]
We denote
\[
\error_p^*=\inf_{f}\E{(f(G_p)-Y)^2}^{\frac{1}{2}},
\]
where the infimum is taken over all measurable functions from \R{p} to
\R{}. As $\E{|Y|^2}<\infty$, $\error_p^*$ is reached for $f$ defined by
$f(g)=\E{Y|G_p=g}$. 

Each function $h$ in $\mathcal{H}_{np}$ can be written $h=f\circ \pi_p$, where
$f$ is chosen in $\mathcal{F}_{np}$ defined by
\[
\begin{split}
    \mathcal{F}_{np}=\Biggl\{&f\in C(\R{p},\R{})\Biggr|
    \\
&
    f(x)=\sum_{l=1}^{L_n}a_lT\left(\beta_{l0}+\sum_{k=1}^p\beta_{lk}x_k\right),
    \mathrm{ with }\sum_{k=1}^{L_n}|a_l|\leq \alpha_n\Biggr\},
\end{split}
\]
Moreover, a function $f_{np}\in \mathcal{F}_{np}$ such that
$h_{np}=f_{np}\circ \pi_p$ 
has obviously the smallest empirical error among functions in
$\mathcal{F}_{np}$, that is
\[
\frac{1}{n}\sum_{i=1}^n(f_{np}(G^i_p)-Y^i)^2\leq \frac{1}{n}\sum_{i=1}^n(f(G^i_p)-Y^i)^2,
\]
for all $f\in \mathcal{F}_{np}$. Then, according to theorem 2 from
\cite{LugosiZeger1995} and thanks to hypothesis on $L_n$ and $\alpha_n$, for
any fixed $p$, $\lim_{n\rightarrow\infty}\error_p(f_{np})=\error_p^*$. In
other words, $\lim_{n\rightarrow\infty}\error_p(h_{np})=\error_p^*$ (almost
surely).

\subsubsection{Step two}
We show now that $\lim_{p\rightarrow\infty}\error_p^*=\error^*$. Let us
consider the sequence of random variables $X_p=\E{Y|G_p}$ and the sequence of
$\sigma$-fields $\mathcal{M}_p=\sigma(G_p)$. We first show that
$(\mathcal{M}_p)_{p\in\N{*}}$ is a filtration, i.e., that
$\mathcal{M}_{p}\subset \mathcal{M}_{p+1}$. This is a simple consequence of
the definition of $G_p$. Indeed, $G_{p}=\pi_p(G)$ and therefore
$G_{p}=\nu_p(G_{p+1})$ where $\nu_p$ is the function from \R{p+1} to \R{p}
defined by
\[
\nu_p(x_1,\ldots,x_p,x_{p+1})=(x_1,\ldots,x_p).
\]
As $G_p$ is the composition of a continuous function and of $G_{p+1}$, the
$\sigma$-field generated by $G_p$ is a subset of the $\sigma$-field generated
by $G_{p+1}$.

As $\E{|Y|^2}<\infty$, $\E{|Y|}<\infty$. This allows to apply Lemma 35 from
\cite{Pollard2002} (page 154), from which we conclude that $(X_p)_{p\in\N{*}}$
is an 
uniformly integrable martingale for the $\mathcal{M}_p$ filtration. Therefore,
according to Theorem 36 from \cite{Pollard2002} (page 154),
$(X_p)_{p\in\N{*}}$ converges almost surely to an integrable random variable
$X_\infty$. Moreover, as $X_p=\E{Y|\mathcal{M}_p}$, according to the same
theorem,
$X_\infty=\E{Y\Bigl|\sigma\left(\bigcup_{p\in\N{*}}\mathcal{M}_p\right)}$.
Obviously, we have
$\sigma\left(\bigcup_{p\in\N{*}}\mathcal{M}_p\right)=\sigma(G)$ and therefore
$(X_p)_{p\in\N{*}}$ converges almost surely to $\E{Y|G}$. 

Finally, as $\E{|Y|^2}<\infty$, $\E{|X_p|^2}\leq\E{|Y|^2}<\infty$ and
therefore, the convergence also happens for the quadratic norm (see Corollary
6.22 from \cite{Kallenberg1997}), i.e.
\[
\lim_{p\rightarrow\infty}\E{\left(\E{Y|G_p}-\E{Y|G}\right)^2}^{\frac{1}{2}}=0.
\]
This clearly implies $\lim_{p\rightarrow\infty}\error_p^*=\error^*$ (almost
surely). 

\begin{acknowledgements}
The authors thank the anonymous referees for their valuable suggestions that
help improving this paper. 
\end{acknowledgements}

\include{fmlp-projection-npl-preprint.bbl}

\end{article}
\end{document}

%% file: fmlp-projection-npl-preprint.bbl.tex
\newcommand{\contenu}[1]{} \newcommand{\dossier}[1]{}